\documentclass{article}

\usepackage{amsmath, amsfonts, amssymb, times, graphicx, natbib, algorithm, algorithmic, hyperref, dsfont, flushend, subcaption}

\newcommand\T{\rule{0pt}{2.6ex}}       % Top strut
\newcommand\B{\rule[-1.2ex]{0pt}{0pt}}
\usepackage[accepted]{data4good2016}

\icmltitlerunning{Predicting Ambulance Demand}

\begin{document}

\twocolumn[
\icmltitle{Predicting Ambulance Demand: Challenges and Methods}

\icmlauthor{Zhengyi Zhou}{zzhou@research.att.com}
\icmladdress{Research done at Cornell University, now at AT\&T Labs Research,
            33 Thomas Street, New York, NY 10007 USA}
%\icmlauthor{Second Author}{sa@us.ibm.com}
%\icmlauthor{Third Author}{ta@us.ibm.com}
%\icmladdress{IBM Thomas J. Watson Research Center,
%            1101 Kitchawan Rd., Yorktown Heights, NY 10598 USA}

\vskip 0.3in
]

\begin{abstract}
Predicting ambulance demand accurately at a fine resolution in time and space (e.g., every hour and  1 km$^2$) is critical for staff / fleet management and dynamic deployment. There are several challenges: though the dataset is typically large-scale, demand per time period and locality is almost always zero. The demand arises from complex urban geography and exhibits complex spatio-temporal patterns, both of which need to captured and exploited. To address these challenges, we  propose three methods based on Gaussian mixture models, kernel density estimation, and kernel warping. These methods provide spatio-temporal predictions for Toronto and Melbourne that are significantly more accurate than the current industry practice.
\end{abstract}

\section{Introduction}\label{intro}

A primary goal of emergency medical services (EMS) is to minimize response times to emergencies while managing operational costs. Sophisticated operations research methods have been developed to optimize many management decisions \citep{Henderson:2009}, but these methods rely critically on accurate, fine-grain demand predictions as inputs. These predictions are crucial to operations decisions such as staff / fleet management, placement of base locations, and dynamic deployment. The industry typically predicts for every hour and every 1-km$^2$ region. 

The current industry practice to predict ambulance demand is crude, while the few methods in prior literature are barely more accurate (details in Section~\ref{lit}). Apart from accuracy, computational speed, robustness, and accessibility to EMS managers are also important considerations.  Methods developed for this problem can also be helpful to police and fire dispatch, or other emergency services.

\subsection{Data and Challenges} \label{data}
We are motivated to predict spatio-temporal ambulance demand for Toronto, Canada and Melbourne, Australia. For Toronto, there were 391,296 priority emergency events in years 2007 and 2008 for which an ambulance was dispatched. Each record contains the time and the location to which the ambulance was dispatched. For Melbourne, we have a similar dataset of 696,975 events for years 2011 and 2012. The two datasets are shown in Figure \ref{fig:data}.

\def\LW{\dimexpr.91\linewidth-.5em}
\def\LWW{\dimexpr.09\linewidth-.5em}
%\vspace{-2mm}
\begin{figure}[!ht]
\centering 
\parbox{\LWW}{(a)}
\parbox{\LW}{\includegraphics[width=\linewidth,height=1.8in]  {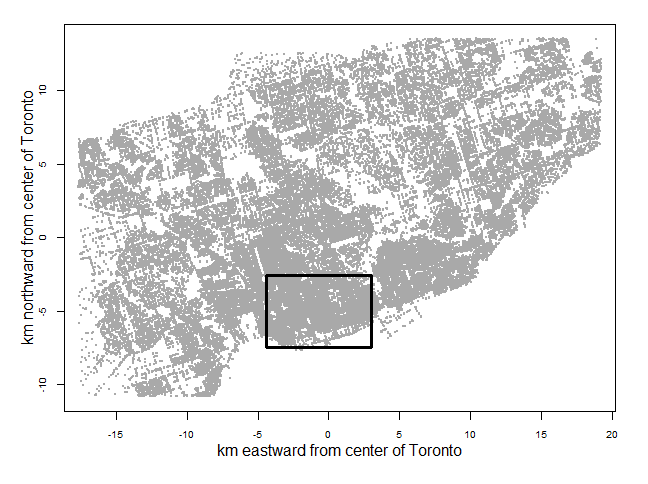}}\\
\parbox{\LWW}{(b)}
\parbox{\LW}{\includegraphics[width=\linewidth,height=1.8in]{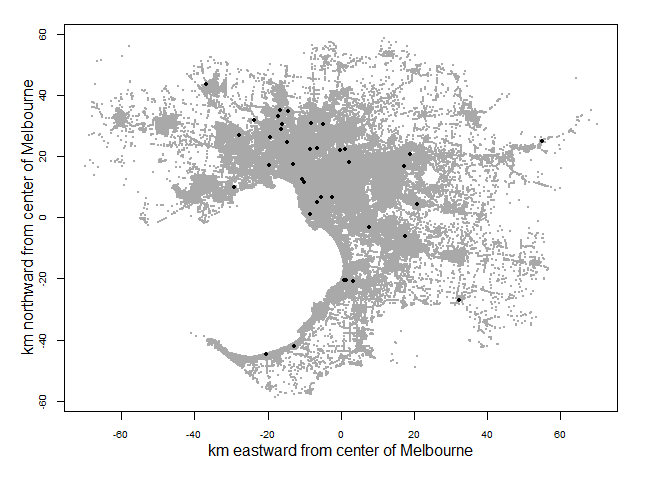}}\\
%\vspace{-4mm}
\caption{(a): Spatial locations of all Toronto ambulance demand incidents in 2007 - 2008, with the downtown region boxed. (b): All 696,975 incidents in Melbourne in 2011 - 2012 (gray) and 37 incidents for a typical 1-hour period (black).}
\label{fig:data}
\end{figure}
%\vspace{-2mm}
Typical challenges to predicting ambulance demand are 
%\vspace{-2mm}
\begin{itemize}
\item The demand is often exceedingly sparse at the temporal and spatial resolution required for prediction. For example, Toronto receives only 23 calls per hour on average; 96\% of the 1-km$^2$ spatial regions have zero calls in any hour. Similarly, Melbourne only receives 37 calls per hour; 99.6\% of the 1-km$^2$ regions receive no calls in any hour (Figure \ref{fig:data}(b)).  
\item There are important spatial and temporal patterns in this data. Weekly seasonality is prominent in both cities; the industry relies heavily on this to make predictions. Temporal patterns also vary by location: in downtown Toronto (boxed in Figure \ref{fig:data}(a)), there are also daily seasonalities and short-term serial dependence that are stronger than other locations in the city.
\item The demand arises from complex urban geography. For example, Melbourne has a highly complex spatial boundary as it encloses a bay to its southwest (Figure~\ref{fig:data}(b)). Demand is high near the bay (coastal downtown), at small suburban neighborhoods, and along major highways radiating out from downtown to suburbs. Demand is low at internal reservoirs and parks. 
\item Ambulance demand data for large cities is often large-scale. Every year, Toronto receives nearly 200,000 calls, and Melbourne receives more than 330,000. This presents computational challenges, especially since predictions are needed very frequently.
\item Data other than historical demand (e.g., weather, demographics) is typically not used in prediction. The most relevant data is how the city's population moves hourly, which is typically not available.
\end{itemize}
%\vspace{-2mm}
It is particularly difficult to simultaneously resolve these challenges. Overcoming sparsity requires considerable smoothing, while capturing complex spatio-temporal patterns requires fine-resolution modeling. At high granularities, data sparsity makes it difficult to detect spatio-temporal characteristics accurately. At low granularities, differences across regions and times are not sufficiently captured for optimal ambulance planning.

\subsection{Current Practice and Related Methods} \label{lit}
The current industry practice for predicting ambulance demand often uses a simple averaging formula. Demand in a 1-km$^2$ spatial region over an hour is typically predicted by averaging a small number of historical counts, from the same spatial region and over the corresponding hours from previous weeks or years. For example, Toronto EMS averages 4 historical counts in the same hour of the year over the past 4 years, while the EMS of Charlotte-Mecklenburg, NC averages 20 historical counts in the same hour of the preceding 4 weeks for the past 5 years (MEDIC method) \cite{Setzler:2009}. Averaging so few historical counts, which are mostly zeros, produces highly noisy and flickering predictions, resulting in haphazard deployment. 

Many studies have accurately predicted the total ambulance demand of a city as a temporal process, using e.g., autoregressive moving average models \cite{Channouf:2007}, and factor models \cite{Matteson:2011}. However, few studies have modeled spatio-temporal ambulance demand well due to data sparsity. Setzler et al. \yrcite{Setzler:2009} use artificial neural networks to predict ambulance demand on discretized time and space, but fails to improve the accuracy over the industry practice.

We propose three new methods to better address this problem: time-varying Gaussian mixture model (\ref{gmm}), spatio-temporal kernel density estimation (\ref{stkde}), and kernel warping (\ref{warp}). We show significant accuracy improvements over the industry practice in Toronto and Melbourne (\ref{result}).

\section{Models} \label{model}
We model ambulance demand on continuous space $\mathcal{S}\subseteq \mathds{R}^2$ and discretized hourly intervals $\mathcal{T}=\{1,2, \ldots, T\}$. Let $\boldsymbol{s}_{t,i}$ denote the spatial location of the $i$th ambulance demand event occurring in the $t$-th time period, for $i \in \{1,\dots, n_t\}$. We assume that the set of spatial locations in each time period independently follows a non-homogeneous Poisson process over $\mathcal{S}$ with an intensity function $\gamma_t$. We further decompose $\gamma_t(\boldsymbol{s}) = \delta_t f_t(\boldsymbol{s})$, where $\delta_t = \int_{\mathcal{S}} \gamma_t(\boldsymbol{s})\, d\boldsymbol{s}$ is the aggregate demand intensity, for modeling total call volume in period $t$, and $f_t(\boldsymbol{s})$ is the spatial \textit{density} of demand in period $t$ that integrates to 1 over $\mathcal{S}$. Thus, $n_t | \gamma_t \sim \mbox{Poisson}(\delta_t)$ and $\boldsymbol{s}_{t,i} | \gamma_t, n_t \sim f_t(\boldsymbol{s})$ iid for $i \in \{1,\ldots,n_t\}$. Many prior studies propose methods for estimating $\{ \delta_t \}$. Here, we focus on estimating $\{f_t(\boldsymbol{s})\}$, which has received little consideration.

\subsection{Time-Varying Gaussian Mixture Model (GMM)} \label{gmm}
In Zhou et al. \yrcite{Zhou:2015a}, we consider an $m$-component Gaussian mixture model in which the component distributions are common in time, while mixture weights change over time. Fixing the component distributions promotes information sharing across time for an accurate spatial structure, overcoming data sparsity within each time period. The component distributions aim to model time-invariant urban structures such as downtown, residential areas, and central traffic routes. The time-varying mixture weights capture dynamics in population movements and actions at different locations and times.  We have
%\vspace{-1mm}
\begin{equation*}
f_t(\boldsymbol{s}) = \sum_{j = 1}^{m}p_{t,j}\,\phi(\boldsymbol{s};\boldsymbol\mu_{j},\boldsymbol\Sigma_{j}), 
\end{equation*}
%\vspace{-4mm}
where $\phi$ is the bivariate Gaussian density, with time-invariant mean $\boldsymbol \mu_j$ and covariance $\boldsymbol\Sigma_j$. The time-varying weights $\{p_{t,j}\}$ are non-negative and sum to one for each $t$. 

Given strong weekly seasonality, we constrain all time periods with the same position within a week (e.g., every Monday 8 - 9am) to have the same mixture weights. That is, for a weekly cycle of $B$ periods, we set  $p_{t,j} = p_{b,j}$ for all $t = b \mbox{ (mod }B)$, $b\in \{1,\ldots,B\}$.

We also observe short-term serial dependence and daily seasonality with varying strengths at different locations. We capture this by placing a separate conditionally autoregressive (CAR) prior on each series of mixture weights $\{p_{b,j}\}_{b=1}^B$ for each component $j$. With such priors, we can represent a rich set of temporal dependence structures, and set unique specification and parameters for each component, allowing us to model location-specific temporal patterns.

Since $\{p_{b,j}\}_{j=1}^m$ for each time $b$ must be non-negative and sum to one, we transform them into unconstrained weights $\{q_{b,r}\}_{r = 1}^{m-1}$ via the multinomial logit transformation, and specify CAR priors on the transformed weights. We assume that the de-meaned $q$ from any time period depend most closely on $q$s from four other periods: immediately before and after (short-term serial dependence) and exactly one day before and after (daily seasonality), i.e., 
%\vspace{-3mm}
\begin{multline*}
q_{b,r} \sim \mbox{N} (a_r+\psi_r[(q_{b-1,r}-a_r)+(q_{b+1,r}-a_r) \\
+(q_{b-d,r}-a_r)+(q_{b+d,r}-a_r)],\nu_r^2),  
\end{multline*}
%\vspace{-4mm}
where $d$ is the number of periods in a day. The CAR parameters $\psi_r$ model the persistence in the weights, while $a_r$ are the means, and $\nu_r^2$ control the variability. These parameters are component-specific, and thus location-specific. 

We use Bayesian estimation, implemented using Markov chain Monte Carlo. See Zhou et al. \yrcite{Zhou:2015a} for computational details, and additional model improvements, including determining the number of components and incorporating weather or demographics data.

\subsection{Spatio-Temporal Kernel Density Estimation (stKDE)} \label{stkde}

In Zhou \& Matteson \yrcite{Zhou:2015b}, we propose stKDE as a method to predict ambulance demand, which is fast, accurate, and easy to interpret and use by non-experts such as EMS managers. First, we parametrically learn the temporal and spatial characteristics of the demand. Each historical demand is annotated with a weight based on what we have learned. This spatio-temporal weight function scores how helpful each historical demand is to a given predictive task. Then we construct a spatial kernel density estimator weighted by the informativeness weight function, and use the resulting kernel density estimates as predictions. In this way, we efficiently emphasize the historical data most important to prediction and, as far as possible, exploit the spatial and temporal characteristics in the data.

We predict the spatial density $f_t$ by aggregating bivariate Gaussian spatial kernel $k$ placed at the location of each past incident $\mathbf{s}_{u, i}$, and weight this kernel by a weight function $w(\mathbf{s}_{u, i}, t)$ that encodes the helpfulness of $\mathbf{s}_{u, i}$ in predicting for the $t$-th time period:
%\vspace{-2mm}
\begin{equation*}
f_{t}(\mathbf{s}) \propto \sum_{\mathbf{s}_{u, i} \in \mathrm{Hist}} w(\mathbf{s}_{u, i}, \, t) \,\, k (\mathbf{s}, \,\mathbf{s}_{u, i}).
\end{equation*}
%\vspace{-4mm}
We aim to incorporate in $w$ the spatial and temporal dependencies in the demand. We discretize $\mathcal{S}$ into 20 large cells, and assume spatially uniform $w$ within each cell. For each cell, we assume the informativeness of a past demand from time $u$ in predicting for future time $t$ only depends on how far back $u$ is from $t$. For cell $c$, we model $w$ as
\begin{equation*}
w(\mathbf{s}_{u,i}, \, t) = \rho_{1,c}^{t-u} + \rho_{2,c}^{t-u}\,\,\rho_{3,c}^{\sin^2\!\left(\!\frac{\pi(t-u)}{T_1}\!\right)\!} \,\,\rho_{4,c}^{\sin^2\!\left(\!\frac{\pi(t-u)}{T_2}\!\right)\!},
\end{equation*}
for $\mathbf{s}_{u,i}$ in cell $c$. Here, the $\rho$s take values in $[0,1]$. The term $\rho_{1,c}^{t-u}$ describes any potential short-term serial dependence. The $\rho_{3,c}$ term models any potential daily seasonality with $T_1 = 24$; the $\rho_{4,c}$ term models weekly seasonality with $T_2= 168$.  These two seasonality terms are multiplied, discounted by $\rho_{2,c}^{t-u}$, and added to the serial dependency effect. The $\rho$ terms are defined and combined similarly to covariance functions in Gaussian processes. We use a separate $\rho$ to capture each typical EMS pattern in each of the 20 cells for easy interpretation, visualization, and comparisons across locations and times, even for non-experts. 

To estimate $\rho$s, we compute the autocorrelation function (ACF) of demand densities for each cell, and find $\rho$s such that $w$ best fits the shape of the positive part of the ACF. This computation is very fast and easily parallelizable. See Zhou \& Matteson \yrcite{Zhou:2015b} for more details, and other model refinements such as interpolating $\rho$ values spatially and omitting less helpful historical data for additional speed.

\begin{figure*}[t]
\centering
	\begin{subfigure}{0.242\linewidth}
	\centering \caption{Toronto: MEDIC} %\vspace{-2mm}
	\includegraphics[width = \linewidth, height = 1.1in]{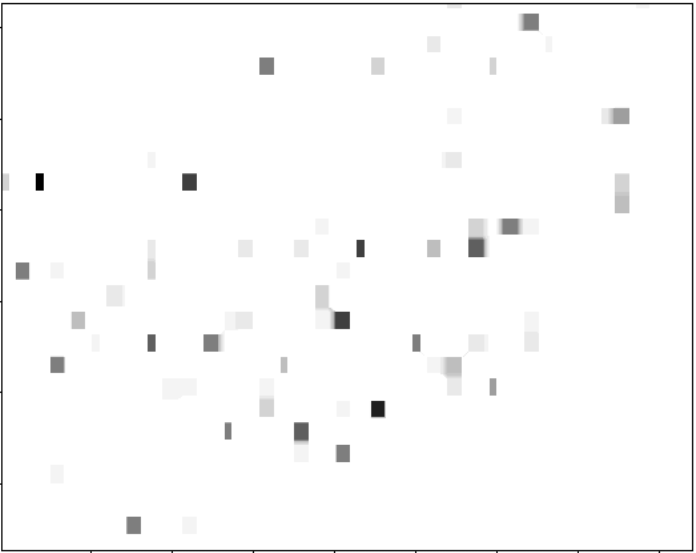}
	\end{subfigure}	
	\begin{subfigure}{0.242\linewidth}
	\centering\caption{naiveKDE}%\vspace{-2mm}
	\includegraphics[width = \linewidth, height = 1.1in]{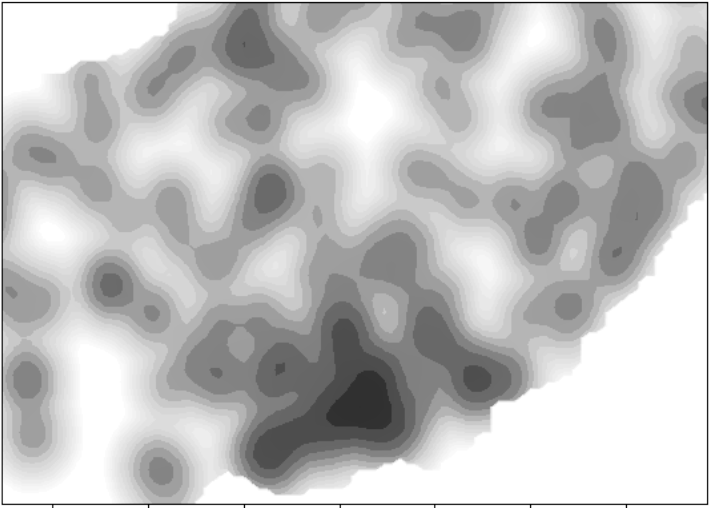}
	\end{subfigure}
	\begin{subfigure}{0.242\linewidth}
	\centering\caption{GMM}%\vspace{-2mm}
	\includegraphics[width = \linewidth, height = 1.1in]{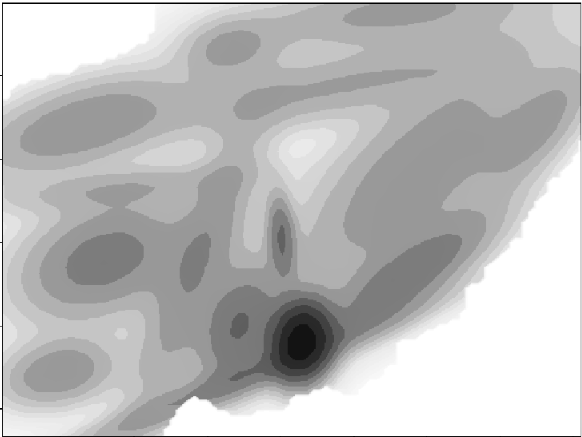}
	\end{subfigure}	
	\begin{subfigure}{0.242\linewidth}
	\centering\caption{stKDE}%\vspace{-2mm}
	\includegraphics[width = \linewidth, height = 1.1in]{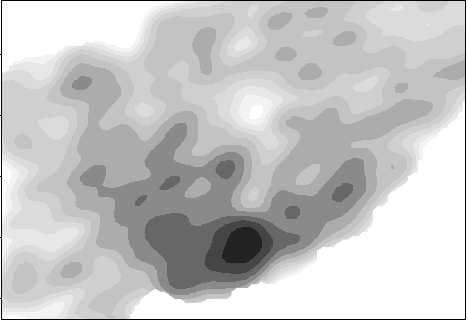}
	\end{subfigure}		
	\begin{subfigure}{0.242\linewidth}
	\vspace{2mm}
	\centering\caption{Melbourne: MEDIC}%\vspace{-2mm}
	\includegraphics[width = \linewidth, height = 1.1in]{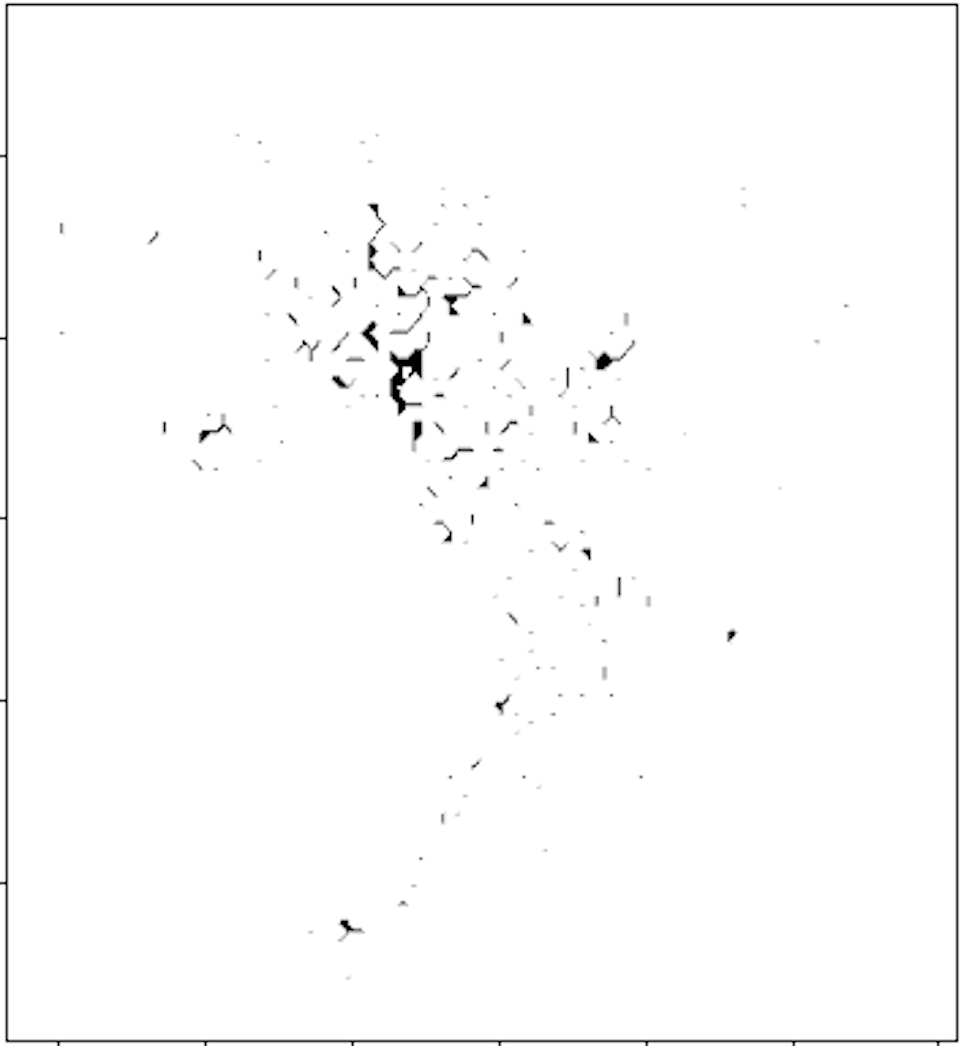}
	\end{subfigure}	
	\begin{subfigure}{0.242\linewidth}
	\centering\caption{naiveKDE}%\vspace{-2mm}
	\includegraphics[width = \linewidth, height = 1.1in]{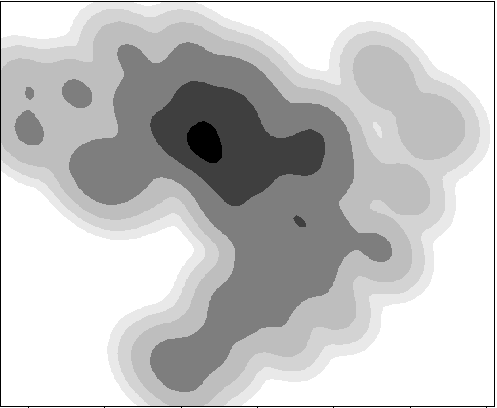}
	\end{subfigure}
	\begin{subfigure}{0.242\linewidth}
	\centering\caption{GMM}%\vspace{-2mm}
	\includegraphics[width = \linewidth, height = 1.1in]{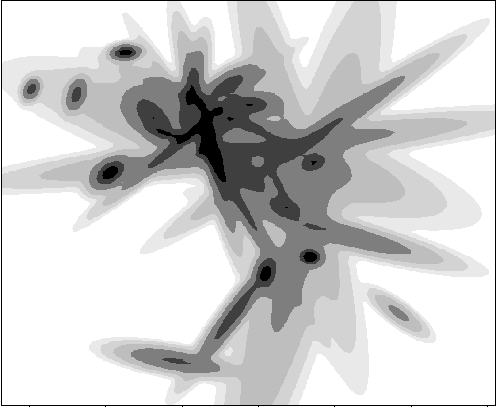}
	\end{subfigure}	
	\begin{subfigure}{0.242\linewidth}
	\centering\caption{WARP}%\vspace{-2mm}
	\includegraphics[width = \linewidth, height = 1.1in]{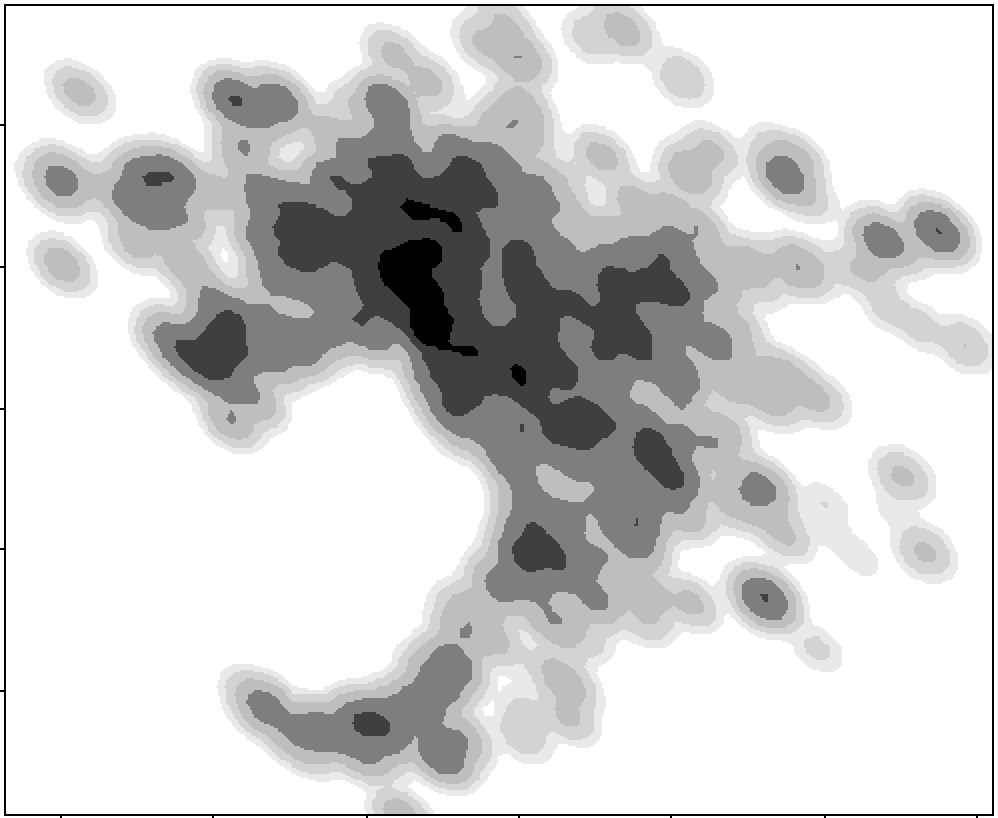}
	\end{subfigure}
	%\vspace{-1mm}
	\caption{Log predictive densities for Toronto (a) - (d)  and Melbourne (e) - (h).}
	\label{fig:preden}
	%\vspace{-2mm}
\end{figure*}

\subsection{Kernel Warping (WARP)} \label{warp}

In Zhou \& Matteson \yrcite{Zhou:2016}, we propose the kernel warping method, mainly motivated by the need to capture a city's complex spatial structures and geographical features. To predict for a future period, we have a sparse set of historical incidents most relevant for this prediction (labeled data). We fit a KDE on them, but warp the kernels to a larger set of historical data regardless of their direct relevance to this predictive task (point cloud). This point cloud describes the spatial structure on which the labeled data lies. It captures exterior and interior boundaries without the need to explicitly define boundaries and boundary conditions. It also incorporates a wide range of complex spatial similarities and discontinuities, such as roads, city blocks, and neighborhoods of varying shapes and densities. Intuitively, this warping can be thought of as a regularization that penalizes radical departure from and encourages information flow along our intuition of the geography. In a Bayesian sense, it can also be thought of as imposing a prior based on how similar or different the point process is across different locations. Such a regularization or prior is especially beneficial when the labeled data is sparse.

Given the prominent weekly seasonality, we choose as our labeled data $\{\mathbf{x}\}$ incidents occurred in the same period of week for the past 8 weeks, and define an unweighted bivariate Gaussian kernel $k$ centered at each $\mathbf{x}$. We choose the point cloud $\{\mathbf{z}\}$ as a random sample of 1000 events in the past 8 weeks. We construct the adjacency matrix $A$ of the point cloud by connecting each node with its 5 closest neighbors, and compute the Laplacian matrix $L = D - A$, where $D$ is the diagonal degree matrix, $D_{i,i} = \sum_j A_{i,j}$.

We warp each $k$ towards the point cloud by
%\vspace{-1mm}
\begin{equation*}
\tilde k(\mathbf{x},\mathbf{s})=k(\mathbf{x},\mathbf{s})-\boldsymbol{k}_\mathbf{x}^T (\boldsymbol{I}+\lambda L\boldsymbol{K})^{-1} \lambda L\boldsymbol{k}_\mathbf{s}, 
\end{equation*}
%\vspace{-3mm}
for any $\mathbf{s}\in\mathcal{S}$, where $\boldsymbol{k}_\mathbf{x}$ and $\boldsymbol{k}_\mathbf{s}$ are vectors of kernels evaluated at $\mathbf{x}$ or $\mathbf{s}$ and $\{\boldsymbol{z}\}$. Matrix $\boldsymbol{K}$ is a matrix of kernels evaluated at all pairs of $\{\boldsymbol{z}\}$, and $\boldsymbol{I}$ is an identity matrix. The parameter $\lambda>0$ represents the degree of deformation. 

We choose $\lambda$ and the kernel bandwidth by cross-validation. See Zhou \& Matteson \yrcite{Zhou:2016} for more computational details, and additional model enhancements that allow for point cloud and warping deformation to vary in time and space. Kernel warping is a variant of the Laplacian eigenmap method in manifold learning \cite{Sindhwani:2005}; we offer practical discussions on applying it to spatio-temporal patterns in  Zhou \& Matteson \yrcite{Zhou:2016}.

\section{Results} \label{result}
We apply GMM and stKDE to the Toronto data, and GMM and WARP to the Melbourne data. For each method, we use 4 - 8  weeks of historical data and predict 4 weeks into future. We compare with an industry practice, MEDIC (\ref{lit}), and an unweighted and unwarped KDE (naiveKDE). 

Figure \ref{fig:preden} shows the predictions using various methods. For Toronto, the competing methods MEDIC (a) and naiveKDE (b) produce noisier predictions than GMM (c) and stKDE (d). Melbourne data is sparser, and naiveKDE (f) over-smooths. The elliptical shapes in GMM (g) do not describe the spatial complexities as well as WARP (h).

Table \ref{tab:result} shows the mean negative log likelihood of test data (logLik, smaller is better). We give a range of logLiks for GMM, stKDE, and WARP, corresponding to various model refinements. These three methods give significantly more accurate predictions than the competing methods. We also show in the papers similar advantages if using the root-mean-squared-error metric, high model goodness-of-fit, and substantial operational benefits via a simulation. 
%\vspace{-4.5mm}
\begin{table} [ht!]
	\centering
		\begin{tabular}{l c  c} 
		\hline
		\textbf{Method}  & \textbf{Toronto logLik} & \textbf{Melbourne logLik}  \T\B \\ \hline
		GMM &  6.07 - 6.15 & 7.87 - 7.96\T\\
		stKDE  & 6.10 - 6.11 &  \\   
		WARP & & 7.53 - 7.56 \B \\ \hline
		MEDIC &  8.64  &  10.11 \T  \\  
		naiveKDE &   6.87  & 8.14 \B \\ \hline
		\end{tabular}
		%\vspace{-1mm}
\caption{Predictive performance (smaller is better) of various proposed and competing models on Toronto and Melbourne data. } 
\label{tab:result}
\end{table}
%\vspace{-5mm}
\section{Conclusions} \label{concl}
Fine-resolution spatio-temporal ambulance demand predictions are critical to
optimal ambulance planning. The EMS industry practice and early studies are simplistic and do not give accurate estimates. We provide three much-needed
and highly accurate methods to predict this demand. The methods overcome data sparsity by intelligently borrowing information across, and represent complex spatial and temporal patterns via priors and weights. Besides that, GMM can incorporate external data, WARP captures complex spatial boundary and features, and stKDE is very fast and easily interpretable and accessible by EMS managers.

The three methods provide a set of generalizable tools suitable to analyze other spatio-temporal point processes that require fine-resolution prediction. Direct applications include modeling crime, fire, and other emergencies that require fast dispatching.

\section*{Acknowledgements} 
The author sincerely thanks Toronto EMS and Ambulance Victoria for sharing their data, as well as the co-authors of the three methods described in this paper: David Matteson, Dawn Woodard, Shane Henderson, and Athanasios Micheas.

\bibliography{d4gbib}
\bibliographystyle{icml2016}

\end{document}